\pdfoutput=1
\documentclass[11pt]{article}

\usepackage[]{emnlp2021}

\usepackage{times}
\usepackage{latexsym}
\usepackage{graphicx}
\usepackage{booktabs}

\usepackage[T1]{fontenc}
\usepackage[utf8]{inputenc}

\usepackage{microtype}

\newcommand{\Shatter}{\textit{Shatter}}
\usepackage{amsmath,amsfonts,bm}

\def\eqref#1{equation~\ref{#1}}
\def\1{\bm{1}}

\def\va{{\bm{a}}}

\def\vq{{\bm{q}}}

\def\vx{{\bm{x}}}
\def\vy{{\bm{y}}}

\def\mA{{\bm{A}}}
\def\mB{{\bm{B}}}

\def\mE{{\bm{E}}}

\def\mK{{\bm{K}}}

\def\mP{{\bm{P}}}
\def\mQ{{\bm{Q}}}
\def\mR{{\bm{R}}}
\def\mS{{\bm{S}}}

\def\mV{{\bm{V}}}
\def\mW{{\bm{W}}}
\def\mX{{\bm{X}}}

\DeclareMathAlphabet{\mathsfit}{\encodingdefault}{\sfdefault}{m}{sl}
\SetMathAlphabet{\mathsfit}{bold}{\encodingdefault}{\sfdefault}{bx}{n}
\newcommand{\tens}[1]{\bm{\mathsfit{#1}}}
\def\tA{{\tens{A}}}
\def\tB{{\tens{B}}}

\def\tK{{\tens{K}}}

\def\tN{{\tens{N}}}

\def\tQ{{\tens{Q}}}

\def\tV{{\tens{V}}}

\def\tX{{\tens{X}}}

\newcommand{\softmax}{\mathrm{softmax}}
\newcommand{\sigmoid}{\sigma}

\DeclareMathOperator{\layernorm}{LayerNormalize}
\DeclareMathOperator{\normalizetwo}{L2-Normalize}

\title{\emph{Shatter}: An Efficient Transformer Encoder with Single-Headed Self-Attention and Relative Sequence Partitioning}

\author{Ran Tian \and Joshua Maynez \and Ankur P. Parikh \\
Google Research Language \\
\texttt{\{tianran, joshuahm, aparikh\}@google.com} \\}

\begin{document}
\maketitle
\begin{abstract}

The highly popular Transformer architecture, based on self-attention, is the foundation of large pretrained models such as BERT, that have become an enduring paradigm in NLP. While powerful, the computational resources and time required to pretrain such models can be prohibitive. In this work, we present an alternative self-attention architecture, {\Shatter}, that more efficiently encodes sequence information by softly partitioning the space of relative positions and applying different value matrices to different parts of the sequence. This mechanism further allows us to simplify the multi-headed attention in Transformer to single-headed. We conduct extensive experiments showing that {\Shatter} achieves better performance than BERT, with pretraining being faster per step (15\% on TPU), converging in fewer steps, and offering considerable memory savings (>50\%). Put together, {\Shatter} can be pretrained on 8 V100 GPUs in 7 days, and match the performance of BERT\textsubscript{Base} -- making the cost of pretraining much more affordable.

%The highly popular Transformer architecture uses multi-head self-attention to encode sequences, and relies on position embeddings to incorporate the sequence order. We present an alternative, which integrates sequence order by applying different value matrices to different parts of the sequence. We show that a coarse partition of the sequence suffices to replace position-wise embeddings, and self-attention scores can be reduced to only one head. At the same model size, our new architecture, named \emph{Shatter}, is $5\%$ faster than BERT on GPU and $15\%$ faster on TPU, converges with less steps in pretraining and performs better when finetuned on downstream tasks. Moreover, our model generalizes better when pretrained with short sequence length and finetuned on tasks that require longer sequence encoding. Put together, one can specifically pretrain {\Shatter} with 8 V100 GPUs in 7 days, and match the performance of BERT\textsubscript{Base} -- making the cost of pretraining affordable for common academic labs.

\end{abstract}

\section{Introduction}

Large pretrained models such as BERT \citep{devlin-etal-2019-bert} have become a mainstay of natural language processing (NLP), enabling the transfer of knowledge learned from large amounts of unsupervised data to downstream tasks with a small number of labeled examples. 
%Typically these approaches involve pretraining a Transformer architecture \citep{vaswani2017attention} on large unlabeled corpora using noisy reconstruction objectives such as Masked Language Modeling \citep{devlin-etal-2019-bert, radford2018improving, song2019mass}. 
Due to the size of both the unsupervised data and model architecture,
%underlying architecture and dataset,
pretraining is computationally very expensive, requiring either TPUs or a substantial number of GPUs \citep{liu2019roberta}, which can make experimentation cumbersome. As a result, many have attempted to improve pretraining efficiency from largely two aspects: the model architecture \citep{ke2020rethinking,shaw2018self,he2020deberta} or pretraining objective \citep{yang2019xlnet,clark2020electra}.

In this work, we focus on developing a more efficient  architecture, inspired by the recent literature of modeling sequence attention with relative position embeddings \citep{shaw2018self}. Similar techniques have been integrated in many Transformer variants \citep{yang2019xlnet,raffel2019t5,he2020deberta,ke2020rethinking}, and as we show in this work, relative position embeddings not only achieve better finetuned performance with fewer pretraining steps (\S\ref{sec:finetune}), but also hold the potential of being pretrained on shorter sequences than used in finetuning, which could considerably reduce the memory requirements and accelerate pretraining.

Unfortunately, training with relative position embeddings is much slower per step: While \citet{shaw2018self} reported $7\%$ more training time on GPUs, we found it $3.9$ times slower than BERT on TPUs\footnote{TPUs are much faster than GPUs in matrix multiplication, but are not optimized for the operations that convert between relative and absolute positions as we explain in \S\ref{sec:background}.} (\S\ref{sec:pretrain_cost}). %During the computation, tensors configured in memory according to relative positions has to be converted to absolute positions and vice versa, which increases the cost especially on TPUs (\S~\ref{...}). 
Consequently, the cost in speed hinders the adoption of this model.
%negates much of the advantages of the model.

In this work, we develop an alternative that captures the advantages of relative position embeddings (i.e.~better finetuning performance with fewer pretraining steps, pretraining on shorter sequence length) without the associated speed overhead. Instead of using (absolute or relative) position embeddings, we propose to incorporate sequence order information by softly partitioning the space of relative positions into parts, which can be implemented by a TPU-efficient operation of multiplying the attention matrix with a constant mask.

%This permits us to multiply the attention matrix with a constant mask to incorporate relative position information, which is a TPU-efficient operation. 

Subsequently, we can replace multi-headed attention with a single-headed variant, by changing the activation function from softmax (i.e.~L1-normalized $\exp$ which favors sparse positions of the top attention score) to L2-normalized sigmoid (which favors dense weighting of positions). This single-headed formulation simplifies the concept of self-attention, and leads to further efficiency gains. 

We exhaustively evaluate our model, \emph{Shatter}, across several datasets: The GLUE Benchmark \citep{wang2018glue}, SQuAD \citep{rajpurkar2016squad}, BoolQ \citep{clark2019boolq} and MultiRC \citep{khashabi2018looking}. We show that our model is 15\% faster per pretraining step than BERT on TPU (\S\ref{sec:pretrain_cost}), and achieves better finetuned performance across the datasets, for both base and large size, with fewer pretraining steps (\S\ref{sec:convergence}). Furthermore, we show that our approach generalizes better to longer sequences, and can pretrain on half the sequence length required in finetuning, which reduces the memory footprint by more than 50\% (\S\ref{sec:seqlength}). Put together, we can pretrain the base-size {\Shatter} on 8 V100 GPUs in 7 days and match or outperform BERT\textsubscript{Base},
%\ankur{while I agree Shatter\textsubscript{Large} outperforms BERT\textsubscript{Large} in accuracy, i am not persuaded this claim is true for Base and GPU, especially on squad and the implementation differences between our version and the public base checkpoint}
significantly reducing the computational requirements for pretraining.

\section{Background}
\label{sec:background}

A Transformer encoder \citep{vaswani2017attention} is a stack of self-attention \citep{cheng-etal-2016-long,parikh-etal-2016-decomposable} and feed-forward layers. In the following, we briefly review the self-attention mechanism and some variants to model relative position. 
Vectors are denoted by lowercase bold letters, matrices by uppercase bold, and tensors of higher ranks by bold sans-serif. We use Numpy indexing\footnote{\url{https://numpy.org/doc/stable/reference/arrays.indexing.html}} for tensors, and matrices are row-major.  %\ankur{I think somewhere we need to define that we show two tensors being multiplied it means batch matrix multiplication}

At each layer $k$, the hidden state of a Transformer is represented by an $l\times d$ matrix $\mX^k$, where $l$ is the sequence length and $d$ the hidden size. For simplicity of notation, we omit the $k$ superscript unless the layer index is critical. The hidden vector $\vx_i=\mX[i,:]$ $(0\leq i<l)$ encodes context around position $i$ in the sequence.

From the hidden state, self-attention calculates the query $\mQ$, key $\mK$, and value $\mV$ matrices as trainable linear transformations of $\mX$: 
\begin{align*}
\mQ & =\mX \mW^{Q},\; \mK=\mX \mW^{K},\; \mV=\mX \mW^{V}
\end{align*}
Then, it uses query and key to compute attention scores and construct a weighted average of the value:
\begin{align*}
\mA &= \softmax \left ( \frac{\mQ \mK^\top}{\sqrt{d}} \right ) \\
\bar{\mX} &= \mA \mV
\end{align*}
Here, $\mA$ is an $l \times l$ matrix, with $\mA[i,j]$ representing the attention score from position $i$ to position $j$. The input to the next layer $\mX^{k+1}$ is computed as $F^k(\bar{\mX}^k, \mX^k)$, where $F^k$ is a feed forward network with layer normalization \citep{ba2016layer}.

\paragraph{Multi-headed Attention} In practice, it performs better to divide the inner dimension of $\mQ$, $\mK$ and $\mV$ into $n$ blocks, and run the self-attention mechanism $n$ times in parallel. This is calculated formally by converting $\mQ$ into a rank $3$ tensor $\tQ$ such that:
\begin{align*}
\tQ[h,:,:]=\mQ[:,\tfrac{h}{n}d:\tfrac{h+1}{n}d]\quad(0\leq h<n)
\end{align*}
and similarly $\mK$, $\mV$ into $\tK$, $\tV$, respectively. Then, multi-headed attention is computed by:
\begin{align}
\tA &= \softmax \left ( \frac{\tQ \tK^\top}{\sqrt{d/n}} \right )
\label{eq:att_tensor}\\
\bar{\tX} &= \tA \tV
\label{eq:ctx_tensor}
\end{align}
where multiplication of tensors means batched matrix multiplication. 
%i.e. $\bar{\tX}[h,:,:] = \tA[h,:,:] \times \tV[h,:,:]$. % Ran: I don't like this notation. I think just saying batched matrix multiplication is enough.  
The tensor $\bar{\tX}$ is converted back to a matrix $\bar{\mX}$ by setting
\begin{align}
\bar{\mX}[:,\tfrac{h}{n}d:\tfrac{h+1}{n}d]=\bar{\tX}[h,:,:]\quad(0\leq h<n).
\label{eq:ctx_matrix}
\end{align}
Here, $\tA$ is a $n\times l\times l$ tensor, where each ``attention head'' $\tA[h,:,:]$ is expected to weight sequence positions from different aspects, and the model more efficiently encodes context information.

\paragraph{Position Embeddings} Self-attention does not recognize sequence order; i.e., a permutation of rows in $\mX$ leads to the same permutation of rows in $\bar{\mX}$. To inject order information, Transformer learns an $l\times d$ matrix $\mP$ as ``position embeddings'', besides the word embeddings $\mE$. Namely, given a word sequence $w_0w_1\ldots$ of length $l$, the input vector $\vx^0_i$ of the first layer is calculated by:
\begin{align}
\vx^0_i=\layernorm(\mE(w_i)+\mP[i]).
\label{eq:pos_emb}
\end{align}
Thus, a permutation of the word sequence $w_0w_1\ldots$ does not lead to a permutation of rows in $\mX^0$, thanks to position embeddings.

\paragraph{Relative Position Embeddings} Although being simple, position embeddings only inject order information to the first layer, and the solution is not shift-invariant (e.g., if we shift $w_0w_1\ldots$ one position to the right, the input sequence $\texttt{<pad>}w_0w_1\ldots$ will be encoded into completely different hidden states, not the shift of hidden states of $w_0w_1\ldots$, which is unintuitive of a sequence encoder). As an alternative, the Relative Position Embeddings (RPE) model \citep{shaw2018self} aims to inject stronger notion of position at each self-attention layer, with shift-invariance. The model learns a $(2l-1)\times d$ matrix $\mR^k$ at each layer $k$, which contributes to both attention scores and values, as below:
\begin{align*}
\mS = \mQ &\mK^\top,\; \mS_\text{rel}[i,j-i] = \mS[i,j] \\
\mK_\text{rel} &= \mR\mW^K,\; \mV_\text{rel} = \mR\mW^V \\
\mA_\text{rel} &= \softmax\left(\frac{\mS_\text{rel}+\mQ \mK_\text{rel}^\top}{\sqrt{d}}\right) \\
\mA[i,j] &= \mA_\text{rel}[i,j-i] \\
\bar{\mX} &= \mA\mV+\mA_\text{rel}\mV_\text{rel}
\end{align*}
Here, $\mS_\text{rel}$ and $\mA_\text{rel}$ are $l\times(2l-1)$ matrices with inner index ranging from $-l+1$ to $l-1$, representing attention to relative positions. Intuitively, the calculation corresponds to using $\vx_j+\mR[j-i]$ to obtain key and value, while using $\vx_i$ to obtain query. The multi-headed case is defined similarly.

In the above calculation, RPE requires conversion between $\mS,\mA$ and $\mS_\text{rel},\mA_\text{rel}$, respectively, which changes the memory configuration of these tensors in the accelerator in a less optimized manner\footnote{For example, a float array is divided into blocks of length 32 on GPU, and the array is always padded and aligned to the boundaries of these blocks. On TPU, the alignment is imposed on 2D blocks of size $8\times128$. Conversion between relative and absolute positions will change the memory alignment.}, and increases computational cost. This is especially a bottleneck on TPUs, making the model difficult to scale up to larger size.

Nevertheless, RPE consistently performs better in our experiments (\S\ref{sec:finetune}), and the model can be naturally extended to longer sequences by copying the learnt embeddings $\mR[-l+1]$ and $\mR[l-1]$ to farther relative positions. It serves as a strong motivation for us to design a fast alternative for relative position modeling.

% ~ when computing the attention.
% \begin{align*}
%  e_{ij} = \frac{\vq_i^T \vk_j + \vq_i^T \va_{j-i}}{\sqrt{d}}
%  \end{align*}
% where $\va_{j-i}$ is a learnt relative position embedding since it is only a function of the distance $j - i$.
% As described in \citet{shaw2018self}, computing the second numerator term requires multiple reshape operations. While the speed cost on GPU is small, reshape is inefficient on TPU\footnote{https://cloud.google.com/tpu/docs/performance-guide#tfreshape}, thus hurting the scalability of relative position embeddings to larger models.

\paragraph{Relative Attention Bias} Several Transformer variants \citep{raffel2019t5,he2020deberta,ke2020rethinking} have simplified the relative position modeling, in addition to the (absolute) position embeddings. For example, instead of Eq.\ref{eq:att_tensor}, the T5 model \citep{raffel2019t5} learns an attention bias $\tB$ according to relative positions:
\begin{align*}
\tB[:,i,j]&=\mW^B[g,:]\quad(b_g\leq j-i<b_{g+1}) \\
\tA &= \softmax \left ( \frac{\tQ \tK^\top}{\sqrt{d/n}}+\tB \right )
\end{align*}
In which, $\tB$ is an $n\times l\times l$ tensor obtained from a trainable $m\times n$ matrix $\mW^B$, with the relative position $j-i$ put into one of $m$ fixed buckets: $b_0=-\infty<\ldots b_g\ldots<b_m=\infty$.
The computation of this Relative Attention Bias (RAB) model is much simpler than RPE, but still adds extra cost. In our experiments, using the T5 default $m=32$, pretraining RAB is $20\%$ more costly than the vanilla Transformer on TPU. Performance is improved on some tasks, but not all (\S\ref{sec:glue}).

%\paragraph{BERT} BERT \citep{devlin-etal-2019-bert} is a Transformer architecture with model parameters $\theta$ trained on large amounts of unsupervised data using a noisy reconstruction objective where for each sequence $\vw = w_1,...,w_l$, we construct a masked version $w_M$ by replacing 15\% of the tokens with a \mask token. The masked language modeling objective is then to attempt to reconstruct original sequence $w$ conditioned on $w_m$ 15\% of the tokens are masked out and predicted by the model with the surrounding context: $\textrm{loss}_{\textrm{MLM}} = - \log p_{\theta}(\vw | \vw_M )$.

%The model is typically trained for ~1 million steps on this objective (termed pretraining). The resulting model is then further trained on small amounts of labeled data for a given task using a supervised objective (termed finetuning). Since our main goal is to improve pretraining efficiency, we evaluate all the baseline architectures as well as our proposed one in the context of this pretrain + finetune paradigm.

\section{Our Approach}
\label{sec:approach}

\begin{figure}[t!]
\centering
\includegraphics[scale=0.55,viewport=0 0 5.6in 5.2in,clip]{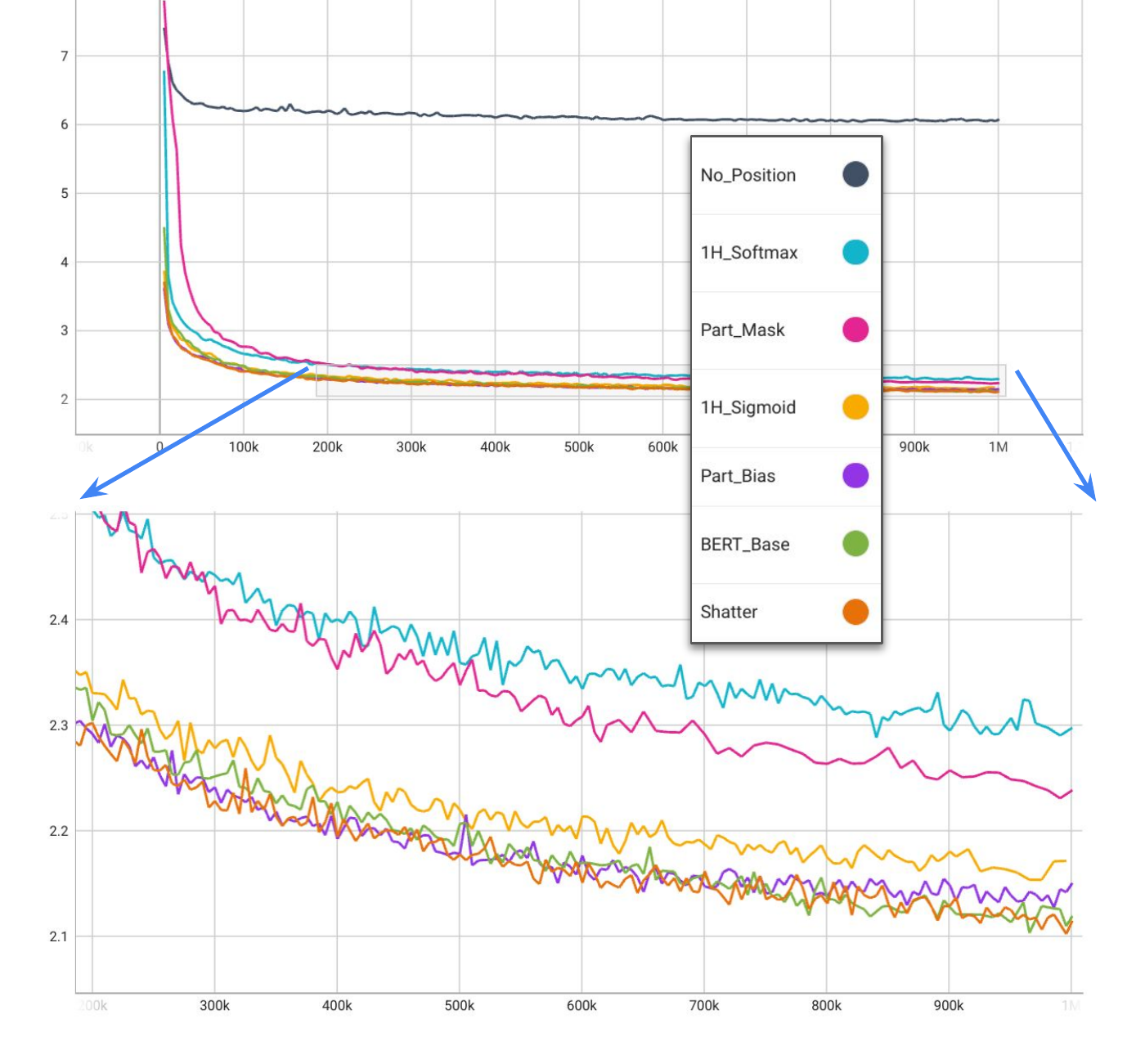}
\caption{The learning curves (MLM valid loss over pretraining step) of our model variants, developed one by one and lead to {\Shatter}, compared with BERT\textsubscript{Base}.}
\label{fig:ablation}
% \vspace{-0.3cm}
\end{figure}

In this section, we describe the components of our more efficient self-attention architecture, {\Shatter}, by steps. First, we partition the space of relative positions to softly bias each attention head to focus on a different part of the sequence (\S\ref{sec:relpartition}). Then, we can simplify the multi-head attention to single-headed (\S\ref{sec:sigmoid-att}). Finally, we show how to incorporate partition embeddings for improved performance with minimal additional cost (~\S\ref{sec:part-embeddings}). 
%\ankur{I added this overview, feel free to restructure}.

A demonstrating experiment is shown in Figure~\ref{fig:ablation}: We progressively develop our model variants % of the same size as BERT\textsubscript{Base} \citep{devlin-etal-2019-bert}, 
and show their Masked Language Modeling (MLM) loss~\cite{devlin-etal-2019-bert} on a validation corpus during pretraining. Detailed settings are given in \S\ref{sec:pretrain_setting}.
The baseline is BERT\textsubscript{Base}, and we note its learning curve converging to $\approx\!2$ (Figure~\ref{fig:ablation}, ``BERT\_Base''). We start from removing the position embeddings in Eq.\ref{eq:pos_emb}, and the valid loss drastically increases to $>\!6$ (Figure~\ref{fig:ablation}, ``No\_Position''), which suggests that the sequence order information is crucial for a Transformer encoder.

% In this section, we present an alternative method of incorporating sequence order into self-attention, which is shift-invariant and can generalize to longer sequences. We describe the method 

\subsection{Soft Relative Partition of Sequence}
\label{sec:relpartition}
Instead of position embeddings, we explore an alternative way of integrating sequence order, by restricting different attention heads to different parts of a sequence. This is achieved by multiplying a mask $\tN$ to the attention score:
\begin{align}
\tA &= \softmax \left ( \frac{\tQ \tK^\top}{\sqrt{d/n}} \right )\odot\tN.
\label{eq:att_mask}
\end{align}
Here, $\odot$ denotes element-wise multiplication, and $\tN$ is an $n\times l\times l$ constant tensor up to model design. Intuitively, the attention head $h$ at position $i$ is more sensitive to position $j$ if $\tN[h,i,j]$ is larger. We set $\tN$ to be dependent only on the relative position $j-i$, which makes the model shift-invariant:
\begin{align*}
\tN[h,i,j]=f_h(j-i).
\end{align*}
In order to recognize sequence order, $f_h$ should not be constant; we use a \textbf{partition of unity} to evenly concentrate different $f_h$ to different parts of the relative position space. 

\begin{figure}[t!]
\centering
\includegraphics[scale=0.51,viewport=0 0 6in 1in,clip]{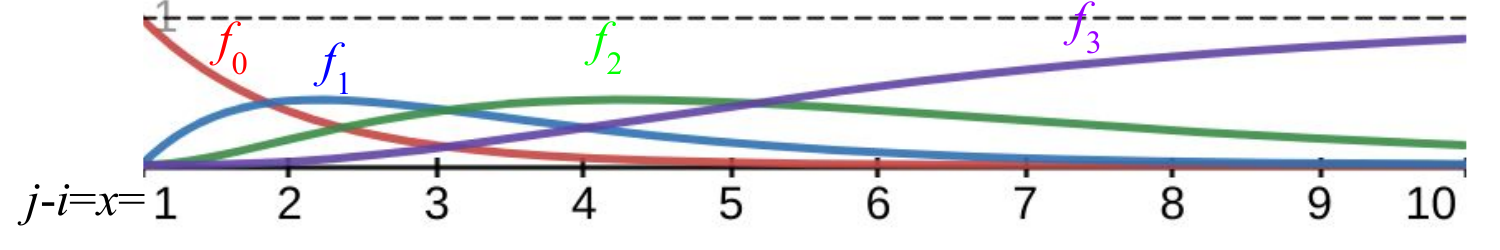}
\caption{An illustration of the partition of unity used in this work, which consists of four functions \{\textcolor{red}{$f_0$}, \textcolor{blue}{$f_1$}, \textcolor{green}{$f_2$}, \textcolor{purple}{$f_3$}\} over the relative position space.
%partition of unity is a group of functions (the red, blue, green and purple graphs) that are non-negative and sum to 1. \ankur{figure needs to be improved i.e. we should make it clear the x-axis is relative positions}
}
\label{fig:part_unity}
% \vspace{-0.3cm}
\end{figure}

A partition of unity is a group of functions $\{f_h\}$ $(0\leq h<n)$ such that: (i) $f_h(x)\geq 0$ and (ii) $\sum_{h=0}^{n-1}f_h(x)=1$, for any $x\in(-\infty, \infty)$. For example, 
%\ankur{not sure if this is true, but i feel we need a terminology for each function, since each function is associated with an embedding later on}
consider the four functions in Figure~\ref{fig:part_unity}: \textcolor{red}{$f_0$} concentrates at relative position $j-i=x=1$ and \textcolor{green}{$f_2$} peaks around $j-i=x=4$, while $f_{0}+f_{1}+f_{2}+f_{3}$ constantly equals $1$ at all positions.

%At $x = 0$ (distance = 0), all the weight is on $f_{red}$, where as $x = ??$, more of the weight is on $f_{green}$. Intuitively, the number of functions will be set to the number of heads $n$, and each will focus the attention on a different part of the (relative) position space \ankur{add more details to this example, as well as to the caption/figure}. 

The partition of unity is a generalization of T5's idea of putting relative positions into buckets: Dividing $(-\infty, \infty)$ into $n$ buckets, $b_0=-\infty<\ldots b_h\ldots<b_n=\infty$, is equivalent to setting 
\begin{align*}
f_h(x)= \begin{cases}
1\quad\mbox{If $b_h\leq x <b_{h+1}$} \\
0\quad\mbox{Otherwise}
\end{cases}.
\end{align*}
%The intuition is that one may not need fine-grained position-wise information to efficiently encode context dependency; a coarse division into $n$ parts might be sufficient, especially for farther positions.
In our case, instead of a hard division into buckets, we adopt a soft division to avoid discontinuity at bucket boundaries. Specifically, we use \textbf{Bernstein polynomials} to construct the partition of unity, and 
%which is illustrated in Figure~\ref{fig:part_unity}. 
the detailed definition is given in \S\ref{sec:part_unity}.

Compared to Eq.\ref{eq:att_tensor}, the extra computation cost of Eq.\ref{eq:att_mask} is negligible, because $\tN$ is constant without any trainable components. Figure~\ref{fig:ablation} (``Part\_Mask'') shows that multiplying $\tN$ largely closes the gap between BERT\textsubscript{Base} and ``No\_Position''.

%\paragraph{Partition Mask:} The partitions are incorporated into the attention computation by multiplying a mask $\tN$ to the attention score:

\subsection{One-Head Sigmoid Attention}
\label{sec:sigmoid-att}

The partition mask $\tN$ makes different attention heads focus on different parts of the sequence, limiting the overlap between parts. It suggests a possibility to neatly combine all the attention scores over the whole sequence into one $l\times l$ matrix, without the multi-headed redundancy. Formally, it suggests that we may replace the tensor $\tQ\tK^\top$ with the matrix version $\mQ\mK^\top$, without loss of model expressivity. For the matrix version, linear algebra suggests that we may further replace $\mK=\mX\mW^K$ by $\mX$, which reduces one matrix multiplication for the self-attention layer. It leads to $5\%$ reduction of training time on GPU and $15\%$ on TPU, in our experiments (\S\ref{sec:pretrain_cost}).

One issue with the matrix version (i.e.~single-headed attention), is that attention scores are calculated by softmax (i.e.~L1-Normalized $\exp$), which favors sparse positions of the top attention score; it might still be better to attend to more positions, as in multi-headed attention. Thus, we propose to use L2-Normalized sigmoid instead of softmax, which favors a dense weighting of positions:
\begin{align}
\tA &= \normalizetwo\left( \sigmoid\left( \frac{\mQ\mX^\top}{\sqrt{d}}\right)\right)\odot\tN.
\label{eq:att_sig}
\end{align}
Figure~\ref{fig:ablation} shows the effect of single-headed attention: If we use softmax scores (``1H\_Softmax''), the validation loss gets slightly worse; however, it is significantly improved by using L2-Normalized sigmoid (``1H\_Sigmoid'').

It is noteworthy that in Eq.\ref{eq:att_sig}, although $\mQ\mX^\top$ is a matrix, we still get an $n\times l\times l$ tensor $\tA$ by multiplying to $\tN$ through Numpy broadcasting. We use Eq.\ref{eq:att_sig} to replace Eq.\ref{eq:att_tensor}, then the tensor $\bar{\tX}=\tA\tV$ is calculated the same as in Eq.\ref{eq:ctx_tensor}. It is crucial to keep $\tV$ as a rank 3 tensor to make the model aware of different parts of the sequence. We have confirmed in preliminary experiments, that reducing $\tV$ to matrix would hurt valid loss, even with sigmoid attention score.

\subsection{Partition Embeddings}
\label{sec:part-embeddings}

Besides injecting sequence order, (absolute or relative) position embeddings enable a model to attend to a position \emph{no matter what the token is}. To achieve a similar role, we further add \textbf{partition embeddings} to our final model, named {\Shatter} (\textbf{S}ingle-\textbf{h}eaded \textbf{atte}ntion with \textbf{r}elative partition).

At each layer, {\Shatter} learns an $n\times d$ matrix $\mR$ to represent partition, in analogue to the relative position embeddings. It uses the query $\mQ$, partition embeddings $\mR$, and the partition mask $\tN$ to calculate an attention bias $\mB$ as the following:
\begin{gather*}
\mB[i,:] =\sum_{h=0}^{n-1}(\mQ\mR^\top)[i,h]\cdot\tN[h,i,:] \\
\tA = \normalizetwo\left( \sigmoid\left( \frac{\mQ\mX^\top}{\sqrt{d}}+\mB\right)\right)\odot\tN
\end{gather*}
In addition, $\mR$ contributes to the value as below:
\begin{gather*}
\mA_\text{part}^\top=\sum_{j=0}^{l-1}\tA[:,:,j],\;\mV_\text{part}=\mR\mW^V \\
\bar{\bar{\mX}} = \bar{\mX} + \mA_\text{part}\mV_\text{part}
\end{gather*}
where $\bar{\mX}$ is defined the same as in Eq.\ref{eq:ctx_matrix}. Then, $\bar{\bar{\mX}}$ is used in the feed-forward network instead of $\bar{\mX}$.% \ankur{$\bar{\tens{X}}$ is not defined here so this is ambiguous and fed-forward is strange word. Can't we just define $\bar{\tens{X}}$ here to be the same as before}

In Figure~\ref{fig:ablation}, ``Part\_Bias'' shows the effect of adding the bias term $\mB$, and ``Shatter'' shows the result adding value contribution. Our final model now matches the valid loss of BERT\textsubscript{Base}.

%Compared to relative position embeddings, 
The computation cost of partition embeddings is small, because $n\ll l$, and the part-wise bias for the attention score is converted to absolute positions via matrix multiplication by a constant tensor $\tN$, without memory rearrangement.

\subsection{Sequence Classification with Relative Position Modeling}
\label{sec:cls_relative}

\begin{table*}[t!]
\footnotesize
\centering
\begin{tabular}{@{}llrcrrrr@{}}
\toprule
Model        & Size                           & \# Params & Hardware     & Batch & Length & \# Steps & Time \\ \midrule
BERT\textsubscript{GPU}     & 12-layers 12-heads 768-hidden  & 84.9M        & GPU V100$\times8$ & 128   & 256    & 1.6M    &  170\,hrs  \\
Shatter\textsubscript{GPU}   & 12-layers 12-parts 768-hidden  & \textbf{78.0M}        & GPU V100$\times8$ & 128   & 256    & 1.6M    &  \textbf{161\,hrs}  \\ \midrule
% BERT\textsubscript{Base}    & 12-layers 12-heads 768-hidden  & 84.9M        & TPUv4 2x2x4  & 256   & 512    & 1M      &    40\,hrs  \\
% Shatter\textsubscript{Base}  & 12-layers 12-parts 768-hidden  & \textbf{78.0M}        & TPUv4 2x2x4  & 256   & 512    & 1M      &   \textbf{34\,hrs}  \\
% BERT-RPE     & 12-layers 12-heads 768-hidden  & 87.3M        & TPUv4 2x2x4  & 256   & 512    & 1M      &  143\,hrs   \\
% BERT-RAB     & 12-layers 12-heads 768-hidden  & 84.9M        & TPUv4 2x2x4  & 256   & 512    & 1M      &  67\,hrs  \\
% XLNet\textsubscript{Base}   & 12-layers 12-heads 768-hidden  & 92.0M        & TPUv4 2x2x4  & 256   & 512    & 1M      &  100\,hrs  \\ \midrule
BERT\textsubscript{Base}    & 12-layers 12-heads 768-hidden  & 84.9M        & TPUv3 4x4  & 256   & 512    & 1M      &    53\,hrs  \\
Shatter\textsubscript{Base}  & 12-layers 12-parts 768-hidden  & \textbf{78.0M}        & TPUv3 4x4  & 256   & 512    & 1M      &   \textbf{45\,hrs}  \\
BERT-RPE     & 12-layers 12-heads 768-hidden  & 87.3M        & TPUv3 4x4  & 256   & 512    & 1M      &  206\,hrs   \\
BERT-RAB     & 12-layers 12-heads 768-hidden  & 84.9M        & TPUv3 4x4  & 256   & 512    & 1M      &  64\,hrs  \\
XLNet\textsubscript{Base}   & 12-layers 12-heads 768-hidden  & 92.0M        & TPUv3 4x4  & 256   & 512    & 1M      &  123\,hrs  \\ \midrule
BERT\textsubscript{Large}   & 24-layers 16-heads 1024-hidden & 151.0M       & TPUv3 8x8    & 1024  & 512    & 1M      &    180\,hrs \\
Shatter\textsubscript{Large} & 24-layers 16-parts 1024-hidden & \textbf{138.6M}       & TPUv3 8x8    & 1024  & 512    & 1M      &  \textbf{151\,hrs}  \\ \bottomrule
\end{tabular}
%\caption{Pretraining speed for a fixed number of steps. {\Shatter}, is faster than BERT across all sizes and platforms, ~4x faster than BERT-RPE and ~2x faster than BERT-RAB, which are alternative techniques for incorporating relative position information (\S\ref{sec:background}) }
\caption{Number of trainable parameters (excluding word embeddings), pretraining settings and pretraining speed.}
\label{tab:pretrain_cost}
\end{table*}

For sequence classification tasks, BERT adds \texttt{[CLS]} and \texttt{[SEP]} tokens to the input sequence(s) and feeds the final hidden state at \texttt{[CLS]} to a classifier. As pointed out by \citet{ke2020rethinking}, this strategy might hurt models using relative position, since the hidden state at \texttt{[CLS]} might be affected by its position relative to other tokens, which is irrelevant to \texttt{[CLS]}'s role of capturing the semantics of the whole sequence(s). Therefore, \citet{ke2020rethinking} propose a mechanism to ``reset'' relative positions regarding \texttt{[CLS]}, but this mechanism is model specific. Related to the issue, \citet{he2020deberta} propose to model relative positions but add absolute position embeddings to the last few layers.

In this work, we adopt a modified strategy for sequence classification, and apply it to all sequence encoders in our experiments for fair comparison. We first encode \texttt{[CLS]}- and \texttt{[SEP]}- augmented input sequence(s) and record the hidden state $\mX^k$ at each layer $(0\leq k\leq L)$. Then, instead of using \texttt{[CLS]}'s final hidden state $\mX^L[0,:]$, we newly start from a learnt vector $\vy^0$ and recalculate a weighted average of the hidden states at each layer:
\begin{align*}
\vq&=\vy\mW^{Q},\; \mK=\mX \mW^{K},\; \mV=\mX \mW^{V} \\
\va&=\softmax\left(\frac{\vq \mK^\top}{\sqrt{d}}\right),\; \bar{\vy}=\va\mV
\end{align*}
Finally, the result $\vy^L$ is used for classification. In the above, the attention mechanism is replaced with either multi-headed or single-headed sigmoid, according to the sequence encoder model.

Our experiments suggest that the strategy does not affect the performance of BERT, but improves others that model relative positions (\S\ref{sec:cls_strategy}).

\section{Related Work}

The original Transformer \citep{vaswani2017attention} uses fixed sinusoid functions to encode positions, justified by the existence of linear transformations between vectors of different positions. This idea is conveyed further by XLNet \citep{yang2019xlnet}, which models relative positions (hence shift-invariant) and can accumulate context information of arbitrary length. Our relative sequence partitioning method differs by directly affecting the attention score, and the coarse partition imposes more regularization than position-wise vectors. We will further compare with XLNet in our evaluation.

BERT \citep{devlin-etal-2019-bert} learns the position embeddings, and \citet{shaw2018self} model relative positions. Many \citep{raffel2019t5,he2020deberta,ke2020rethinking,wennberg2021case,chen2021demystify} combine the techniques with different approaches, while we point out that modeling relative positions with learnt components often increases pretraining cost, and will clearlly evaluate the cost-performance trade-off.

Many ideas are explored to reduce the training cost \citep{lan2019albert,jiang2020convbert,chen2020earlybert,thorp2021fnet}, some specifically for very long ($\geq\!4096$) sequences \citep{child2019sparse,kitaev2020reformer,choromanski2020performer,beltagy2020longformer}, but we are not aware of previous work that can pretrain with 8 V100 GPUs within 1 week and match BERT\textsubscript{Base}.

Previous works have found that multiple attention heads are redundant and can be pruned \citep{michel2019sixteen,voita2019analyzing}, but incorporating this insight into training is more challenging, as naively reducing the number of heads will reduce performance. Our single-headed self-attention both simplified the concept and improved performance.

%Concurrent to our work,  \citet{wennberg2021case} propose a technique that adds a relative position matrix to the attention matrix. While their approach leads to improvements when supplementing normal position embeddings, it leads to performance degradation if normal position embeddings are removed. On the other hand, our approach only requires an embedding for each partition which is much smaller than the number of positions.  \citet{chen2020earlybert} propose a complementary approach to speed up BERT pretraining using the lottery ticket hypothesis.

%Our method is different from the strategy of using fixed  . In our case, the partition of unity is used to directly influence the attention score to attend to different parts of the sequence. 

%\newpage

\section{Experiments}

\begin{table*}[t!]
\scriptsize
\centering
\setlength{\tabcolsep}{2.5pt}
\begin{tabular}{@{}lrrrrrrrrrrrrrrrrrr@{}}
\toprule
                          & \multicolumn{2}{c}{CoLA}                                         & \multicolumn{2}{c}{SST-2}                                         & \multicolumn{2}{c}{MRPC}                                      & \multicolumn{2}{c}{QQP}                                       & \multicolumn{2}{c}{STS-B}                                             & \multicolumn{2}{c}{MNLI-m}                                        & \multicolumn{2}{c}{MNLI-mm}                                       & \multicolumn{2}{c}{QNLI}                                          & \multicolumn{2}{c}{RTE}                                           \\ \cmidrule(rl){2-3} \cmidrule(rl){4-5} \cmidrule(rl){6-7} \cmidrule(rl){8-9} \cmidrule(rl){10-11} \cmidrule(rl){12-13} \cmidrule(rl){14-15} \cmidrule(rl){16-17} \cmidrule(rl){18-19} 
                          & \multicolumn{1}{c}{\begin{tabular}[c]{@{}c@{}}Dev.\\ (Acc.)\end{tabular}} & \multicolumn{1}{c}{\begin{tabular}[c]{@{}c@{}}Test\\ (MCC)\end{tabular}} & \multicolumn{1}{c}{\begin{tabular}[c]{@{}c@{}}Dev.\\ (Acc.)\end{tabular}} & \multicolumn{1}{c}{\begin{tabular}[c]{@{}c@{}}Test\\ (Acc.)\end{tabular}} & \multicolumn{1}{c}{\begin{tabular}[c]{@{}c@{}}Dev.\\ (F1)\end{tabular}} & \multicolumn{1}{c}{\begin{tabular}[c]{@{}c@{}}Test\\ (F1)\end{tabular}} & \multicolumn{1}{c}{\begin{tabular}[c]{@{}c@{}}Dev.\\ (F1)\end{tabular}} & \multicolumn{1}{c}{\begin{tabular}[c]{@{}c@{}}Test\\ (F1)\end{tabular}} & \multicolumn{1}{c}{\begin{tabular}[c]{@{}c@{}}Dev.\\ ($\rho$)\end{tabular}} & \multicolumn{1}{c}{\begin{tabular}[c]{@{}c@{}}Test\\ ($\rho$)\end{tabular}} & \multicolumn{1}{c}{\begin{tabular}[c]{@{}c@{}}Dev.\\ (Acc.)\end{tabular}} & \multicolumn{1}{c}{\begin{tabular}[c]{@{}c@{}}Test\\ (Acc.)\end{tabular}} & \multicolumn{1}{c}{\begin{tabular}[c]{@{}c@{}}Dev.\\ (Acc.)\end{tabular}} & \multicolumn{1}{c}{\begin{tabular}[c]{@{}c@{}}Test\\ (Acc.)\end{tabular}} & \multicolumn{1}{c}{\begin{tabular}[c]{@{}c@{}}Dev.\\ (Acc.)\end{tabular}} & \multicolumn{1}{c}{\begin{tabular}[c]{@{}c@{}}Test\\ (Acc.)\end{tabular}} & \multicolumn{1}{c}{\begin{tabular}[c]{@{}c@{}}Dev.\\ (Acc.)\end{tabular}} & \multicolumn{1}{c}{\begin{tabular}[c]{@{}c@{}}Test\\ (Acc.)\end{tabular}} \\ \midrule
BERT\textsubscript{GPU} 256-$l$          & 82.0                            & -                              & 92.5                            & -                               & 90.5                          & -                             & 87.2                          & -                             & 84.5                              & -                                 & 83.5                            & -                               & 84.2                            & -                               & 90.8                            & -                               & \textbf{66.4}                   & -                               \\
BERT\textsubscript{GPU} 512-$l$ ext.  & 82.1                            & -                              & \textbf{92.8}                   & -                               & 86.1                          & -                             & 86.7                          & -                             & 84.5                              & -                                 & 82.9                            & -                               & 83.2                            & -                               & 90.1                            & -                               & 63.9                            & -                               \\
Shatter\textsubscript{GPU} 512-$l$ ext.  & \textbf{83.3}                   & 49.1                           & 92.2                            & 92.5                            & \textbf{91.3}                 & 88.7                          & \textbf{87.8}                 & 70.6                          & \textbf{85.5}                     & 85.1                              & \textbf{84.5}                   & 84.2                            & \textbf{84.8}                   & 83.6                            & \textbf{91.4}                   & 90.4                            & 66.1                            & 65.4                            \\ \midrule
BERT\textsubscript{Base}                 & 83.6                            & 46.9                           & 92.2                            & 92.3                            & 89.7                          & 87.6                          & 87.3                          & 70.2                          & 85.6                              & 83.8                              & 84.0                            & 83.5                            & 83.8                            & 82.4                            & 90.7                            & 89.8                            & 65.3                            & 61.8                            \\
Shatter\textsubscript{Base}               & 84.0                            & 48.8                           & \textbf{93.6}                   & 92.8                            & \textbf{92.1}                 & 88.1                          & 87.6                          & 70.5                          & 85.8                              & 81.6                              & 84.8                            & 84.7                            & 84.9                            & 83.7                            & 90.7                            & 90.9                            & 67.9                            & 66.4                            \\
BERT-RPE                  & \textbf{84.4}                   & \textbf{55.7}                  & 93.5                            & \textbf{94.0}                   & \textbf{92.1}                 & \textbf{89.5}                 & \textbf{88.5}                 & \textbf{71.0}                 & 86.4                     & \textbf{85.0}                     & \textbf{86.2}                   & \textbf{86.1}                   & \textbf{86.1}                   & \textbf{85.0}                   & \textbf{92.3}                   & \textbf{92.1}                   & \textbf{74.0}                   & \textbf{70.4}                   \\
BERT-RAB                  & 76.2                            & 33.4                           & 92.4                            & 92.7                            & 88.3                          & 86.3                          & 86.7                          & 69.1                          & 84.3                              & 80.2                              & 85.1                            & 84.4                            & 85.0                            & 84.2                            & 91.2                            & 90.6                            & 65.0                            & 61.7                            \\
XLNet\textsubscript{Base}                & 79.4                            & -                              & 93.0                            & -                               & 90.8                          & -                             & 86.4                          & -                             & 81.8                              & -                                 & 84.1                            & -                               & 84.6                            & -                               & 89.6                            & -                               & 68.2                            & -                               \\
\emph{Huggingface} BERT    & 83.3                           & -                              & 92.4                           & -                               & 91.2                          & -                             & 88.1                        & -                             & \textbf{86.7}                            & -                                 & 84.9                           & -                               & 84.7                          & -                               & 91.5                           & -                               & 70.8                          & -                               \\ \midrule
BERT\textsubscript{Large}                & 84.9                            & 56.5                           & 95.4                            & 94.2                            & 92.4                          & 90.1                          & 88.4                          & 71.6                          & 87.3                              & 85.9                              & \textbf{88.9}                   & 88.2                            & \textbf{88.8}                   & 87.4                            & 93.6                            & 93.0                            & 80.9                            & 75.4                            \\
Shatter\textsubscript{Large}              & \textbf{85.9}                   & \textbf{65.2}                  & \textbf{96.2}                   & \textbf{95.6}                   & \textbf{93.3}                 & \textbf{90.4}                 & 88.7                 & 71.2                          & \textbf{87.9}                     & \textbf{90.3}                     & 88.7                            & \textbf{88.4}                   & 88.6                            & \textbf{87.6}                   & \textbf{94.1}                   & \textbf{93.3}                   & \textbf{84.1}                   & \textbf{77.0}                   \\
BERT\textsubscript{Large} 400k           & 81.2                                                                       & 44.5                                                                      & 93.9                                                                      & 94.7                                                                       & 91.3                                                                     & 84.7                                                                     & 88.3                                                                    & 71.5                                                                     & 86.9                                                                       & 85.6                                                                         & 87.6                                                                       & 87.2                                                                       & 87.6                                                                       & 86.2                                                                       & 93.2                                                                       & 92.5                                                                      & 70.0                                                                       & 63.5                                                                       \\
Shatter\textsubscript{Large} 400k         & 85.5                                                                       & 59.8                                                                      & 95.0                                                                     & 94.1                                                                       & 90.3                                                                    & 88.9                                                                     & \textbf{88.8}                                                        & 71.5                                                                     & 87.3                                                                       & 89.0                                                                         & 87.6                                                                      & 87.5                                                                       & 87.7                                                                      & 86.6                                                                      & 93.6                                                                     & 93.0                                                                       & 70.0                                                                     & 66.7                                                                       \\
\emph{Jacob Devlin} Subm. & -                               & 60.5                           & -                               & 94.9                            & -                             & 89.3                          & -                             & \textbf{72.1}                 & -                                 & 86.5                              & -                               & 86.7                            & -                               & 85.9                            & -                               & 92.7                            & -                               & 70.1                            \\ \bottomrule
\end{tabular}
%\caption{Results on the GLUE benchmark~\citep{wang2018glue}. \emph{Top}: Results for GPU models (see Section~\ref{sec:seqlength}). \emph{Middle}: Results for base-sized models along with several baselines. \Shatter performs favorably to BERT and XLNet\textsubscript{Base}. While BERT-RPE and BERT-RAB perform better they take significantly longer to train (Table~\ref{tab:pretrain_cost}). \emph{Bottom}: Results for large size models showing \Shatter \, consistently outperforms BERT.}
\caption{Results on GLUE. Test performance obtained by submitting to \url{https://gluebenchmark.com/}.}
\label{tab:glue}
\end{table*}

%We next evaluate \Shatter \, by implementing and comparing the following models. 
%\ankur{the length and extension terminology needs to be rigorously defined somewhere}
For evaluation, we implemented and compared the following models:

\vspace{1ex}
\noindent
\textbf{BERT} \citep{devlin-etal-2019-bert} is a Transformer encoder \citep{vaswani2017attention} with model parameters trained on large amounts of unsupervised data (i.e.~pretraining). The Masked Language Modeling (MLM) pretraining objective attempts to recover the original tokens when 15\% tokens in an input sequence are masked out. We pretrain three models, BERT\textsubscript{GPU}, BERT\textsubscript{Base} and BERT\textsubscript{Large}, using different model size (Base or Large), sequence length (256-$l$ or 512-$l$) and hardware (GPU or TPU).

%\paragraph{BERT} BERT \citep{devlin-etal-2019-bert} is a Transformer architecture with model parameters $\theta$ trained on large amounts of unsupervised data using a noisy reconstruction objective where for each sequence $\vw = w_1,...,w_l$, we construct a masked version $w_M$ by replacing 15\% of the tokens with a \mask token. The masked language modeling objective is then to attempt to reconstruct original sequence $w$ conditioned on $w_m$ 15\% of the tokens are masked out and predicted by the model with the surrounding context: $\textrm{loss}_{\textrm{MLM}} = - \log p_{\theta}(\vw | \vw_M )$.

%The model is typically trained for ~1 million steps on this objective (termed pretraining). The resulting model is then further trained on small amounts of labeled data for a given task using a supervised objective (termed finetuning). Since our main goal is to improve pretraining efficiency, we evaluate all the baseline architectures as well as our proposed one in the context of this pretrain + finetune paradigm.

\vspace{1ex}
\noindent
\textbf{Shatter} is our proposed model architecture. We pretrain Shatter\textsubscript{GPU}, Shatter\textsubscript{Base} and Shatter\textsubscript{Large}, in correspondence to the BERT models.

\vspace{1ex}
\noindent
\textbf{BERT-RPE} \citep{shaw2018self} uses Relative Position Embeddings (RPE)  at each layer, instead of (absolute) position embeddings. We set the vocab size of RPE to $2\times 128 - 1=255$, so that relative positions farther than $128$ are represented by the same embedding vector.

\vspace{1ex}
\noindent
\textbf{BERT-RAB} adds a Relative Attention Bias (RAB) to the BERT model. Relative positions are put into buckets, and each bucket shares the same bias term. We copied code from the T5 model \citep{raffel2019t5} to implement bucketing, and the number of buckets is set to $32$ (T5 default).

\vspace{1ex}
\noindent
\textbf{XLNet} \citep{yang2019xlnet} uses fixed sinusoid encoding to represent relative positions, and has a memorization mechanism to accumulate context information of arbitrary length. It is also pretrained by an auto-regressive Permutation Language Modeling (PLM) objective instead of MLM. In this work, we pretrain XLNet with PLM but without memorization, in order to adopt the same input data pipeline as other models.

%\paragraph{Sequence Length Extension:} \ankur{I feel we need some paragraph to discuss notation like this, not sure where to put it} One advantage of relative positions is that it is possible to increase the sequence length in finetuning. In particular, we experiment with \emph{doubling} the sequence length in finetuning which we denote by \emph{ext}. For example, \Shatter \, 512 ext refers to a model trained on length 256 but extended to 512 in finetuning. %It is possible to extend BERT in finetuning by adding extra parameters, which we denote similarly e.g. BERT 512-\ell ext.

All the models were implemented by modifying the code of \emph{Huggingface} Transformers package \citep{wolf-etal-2020-transformers}, using TensorFlow 2\footnote{\url{https://www.tensorflow.org/overview}}. We use the BooksCorpus~\citep{zhu2015aligning} and English Wikipedia to pretrain the models, the same as \citet{devlin-etal-2019-bert} (but not the same as RoBERTa \citep{liu2019roberta}). More detailed settings for pretraining are given in \S\ref{sec:pretrain_setting}.

\subsection{Pretraining Cost}
\label{sec:pretrain_cost}

Table~\ref{tab:pretrain_cost} shows the number of trainable parameters, and the time required to pretrain the models on different hardware. {\Shatter} has a significant reduction in number of parameters than other models, due to the omission of a $d\times d$ matrix per layer (\S\ref{sec:sigmoid-att}). It also leads to 5\% faster pretraining speed than BERT on GPUs and 15\% on TPUs. 

Reducing the batch size to 128 and sequence length to 256 (the largest possible to fit in memory), we can pretrain {\Shatter} on 8 V100 GPUs for 1.6M steps within 1 week. At step 1.6M, Shatter\textsubscript{GPU} sees the same amount of tokens as Shatter\textsubscript{Base} at step 400k; as we will show in \S\ref{sec:glue}, it is already close to converge and outperforms BERT\textsubscript{Base} on GLUE. %(We  also train a sequence length 256, batch size 128 BERT on GPU as a comparison).

BERT-RPE has significantly more parameters than BERT\textsubscript{Base}, due to the relative position embeddings at each layer; which prevents the model from running on GPUs using the same settings as BERT\textsubscript{GPU}. It is also $3.9$ times slower than BERT on TPUs, making it impractical to scale up to larger size.
BERT-RAB has almost the same number of parameters as BERT, but still 20\% slower on TPU; The most memory intensive model is XLNet, due to an extra trainable $d\times d$ matrix at each layer to transform the sinusoid encoding of relative positions. Pretraining of XLNet\textsubscript{Base} is also slow, probably due to the two-stream self-attention calculation for the PLM objective. %\ankur{what does PLM stand for}

%In order to evaluate the quality of pretrained models, we finetune the models on GLUE~\citep{wang2018glue}, SQuAD v1.1~\citep{rajpurkar2016squad}, and MultiRC~\citep{khashabi2018looking} and BoolQ~\citep{clark2019boolq} from SuperGLUE \citep{wang2019superglue}. For all tasks, we use the Adam optimizer \citep{kingma2014adam}, set batch size to 16, evaluate the model in every 1k steps and save the best according to dev.

\subsection{Finetuning Results}
\label{sec:finetune}

%In order to evaluate the quality of pretrained models, 
We finetune the models on GLUE~\citep{wang2018glue}, SQuAD v1.1~\citep{rajpurkar2016squad}, and MultiRC~\citep{khashabi2018looking} and BoolQ~\citep{clark2019boolq} from SuperGLUE \citep{wang2019superglue}. For all tasks, we use the Adam optimizer \citep{kingma2014adam}, set batch size to 16, evaluate the model in every 1k steps and choose the one that scores highest on the development set.

\subsubsection{GLUE Benchmark}
\label{sec:glue}

\begin{figure*}[t!]
\centering
\includegraphics[scale=0.65,viewport=0 0 9.5in 1.6in,clip]{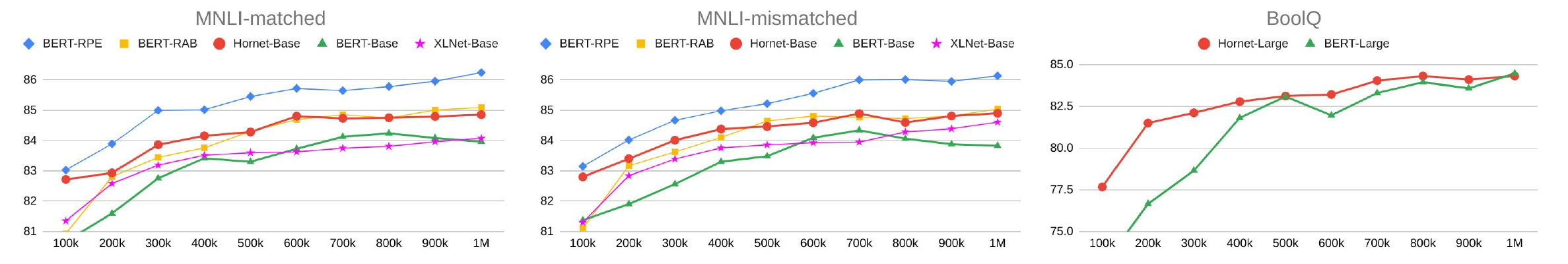}
\caption{Plots showing finetuned performance on dev set (y-axis) as a function of pretrained steps (x-axis).}
\label{fig:converge}
\end{figure*}

GLUE is a collection of classification tasks on text sequences or sequence pairs. Following \citet{devlin-etal-2019-bert}, we evaluate on CoLA~\citep{warstadt-etal-2019-neural}, SST-2~\citep{socher-etal-2013-recursive}, MRPC~\citep{dolan2005automatically}, QQP\footnote{\url{https://quoradata.quora.com/First-Quora-Dataset-Release-Question-Pairs}}, STS-B~\citep{cer2017semeval}, MNLI~\citep{williams-etal-2018-broad}, QNLI~\citep{wang2018glue} and RTE~\citep{bentivogli2009fifth} in GLUE. Table~\ref{tab:glue} shows the results.

For BERT\textsubscript{Base}, \citet{devlin-etal-2019-bert} finetuned with learning rate [1e-5, 2e-5, 3e-5, 4e-5, 5e-5] and took the best. We found smaller learning rates almost always give better results, but the numbers on development set can randomly vary $\sim\!0.25$ points on large datasets such as MNLI or QNLI, and up to $2$ points on small datasets such as RTE or MRPC. Thus, we set the learning rate on CoLA and SST-2 to 5e-6, otherwise to 1e-5, and we conduct several runs to show one run better than average (i.e., if the number on some task is worse than average, we will re-run it and show a better one).

It is especially important to use the small learning rate 5e-6 on CoLA; our accuracy on CoLA dev is better than \citet{devlin-etal-2019-bert} and other papers. On the other hand, our BERT\textsubscript{Base} is $\sim\!1$ point worse on MNLI than \citet{devlin-etal-2019-bert}, which agrees with the findings of a reproduction study \citep{sellam2021multiberts}, that it usually requires BERT\textsubscript{Base} to be pretrained for a larger number of steps in order to reproduce the performance on GLUE. As a reference, we also show the finetuned results of the ``bert-base-uncased'' checkpoint in the \emph{Huggingface} Transformers package (row ``\emph{Huggingface} BERT'' in Table~\ref{tab:glue}), which reproduces \citet{devlin-etal-2019-bert}.

Overall, Shatter\textsubscript{GPU} and Shatter\textsubscript{Base} match \emph{Huggingface} BERT on the development set, and outperform BERT\textsubscript{Base} on test. It is noteworthy that, although Shatter\textsubscript{Base} is better than Shatter\textsubscript{GPU} on dev, the performance gap on test set is only marginal. It suggests that Shatter\textsubscript{GPU} is close to convergence, making the model favorable given its low pretraining cost. Shatter\textsubscript{GPU} achieves such cost-performance thanks to two ingredients: a shorter sequence length in pretraining and a faster convergence of the model. We will further investigate these two aspects in following sections.

BERT-RPE is the best performing among base-size models, demonstrating the efficiency of relative position modeling. However, its high training cost on TPU becomes a burden, as one can outperform BERT-RPE with larger models. On the other hand, BERT-RAB and XLNet\textsubscript{Base} ourperform BERT\textsubscript{Base} on some tasks, but not all.

Scaling up, Shatter\textsubscript{Large} consistently outperforms BERT\textsubscript{Large}, showing the ability of our model. We also cite \emph{Jacob Devlin}'s original BERT\textsubscript{Large} submission to GLUE leaderboard for better comparison.

\subsubsection{Extending Max Sequence Length}
\label{sec:seqlength}
\begin{table}[t!]
\footnotesize
\centering
\setlength{\tabcolsep}{10pt}
\begin{tabular}{lrr}
\toprule
SQuADv1.1 (Dev.)         & \multicolumn{1}{c}{EM} & \multicolumn{1}{c}{F1} \\ \midrule
BERT\textsubscript{GPU} 256-$l$         & 80.0                   & 75.6                   \\
BERT\textsubscript{GPU} 512-$l$ ext.        & 79.7                   & 87.3                   \\
Shatter\textsubscript{GPU} 256-$l$    & 78.8                   & 75.5                   \\
Shatter\textsubscript{GPU} 512-$l$ ext.   & \textbf{81.8}          & \textbf{88.7}          \\ \midrule
BERT\textsubscript{Base} 512-$l$        & 83.3                   & 90.0                   \\
BERT\textsubscript{Base} 1024-$l$ ext.      & 79.7                   & 86.6                   \\
Shatter\textsubscript{Base} 512-$l$   & 82.6                   & 89.6                   \\
Shatter\textsubscript{Base} 1024-$l$ ext. & 82.8                   & 89.6                   \\
BERT-RPE 512-$l$         & \textbf{84.7}          & \textbf{91.0}          \\
BERT-RPE 1024-$l$ ext.       & 84.4                   & \textbf{91.0}          \\ \midrule
BERT\textsubscript{Large} 512-$l$       & \textbf{86.5}          & 92.8                   \\
BERT\textsubscript{Large} 1024-$l$ ext.     & \textbf{86.5}          & 92.7                   \\
Shatter\textsubscript{Large} 512-$l$  & \textbf{86.5}          & 92.8                   \\
Shatter\textsubscript{Large} 1024-$l$ ext. & 86.3                   & \textbf{93.0}          \\ \bottomrule
\end{tabular}
\caption{Results on SQuAD v1.1 development set.}
\label{tab:squad}
\end{table}

\begin{table}[t!]
\footnotesize
\centering
\setlength{\tabcolsep}{10pt}
\begin{tabular}{lrr}
\toprule
MultiRC             & \multicolumn{1}{c}{\begin{tabular}[c]{@{}c@{}}Dev.\\ (Acc.)\end{tabular}} & \multicolumn{1}{c}{\begin{tabular}[c]{@{}c@{}}Test\\ (F1a)\end{tabular}} \\ \midrule
BERT\textsubscript{Large} 512-$l$  & 81.0 &  76.6                     \\
BERT\textsubscript{Large} 1024-$l$ ext.   &  79.6   & 75.8         \\
Shatter\textsubscript{Large} 512-$l$  & 79.4 &  77.1                   \\
Shatter\textsubscript{Large} 1024-$l$ ext. &  \textbf{81.7}  &  \textbf{78.9}   \\ \bottomrule
\end{tabular}
\caption{Results on the MultiRC task.}%~\citep{khashabi2018looking}. All approaches were pretrained on sequence length 512, and \textit{ext.} indicates the model was extended to length 1024 in finetuning. \Shatter can easily adapt to a longer sequence length in finetuning and thus outperforms BERT.}
\label{tab:multirc}
\end{table}

The relative sequence partition in {\Shatter} can be naturally extended longer. In order to demonstrate our model's ability of generalizing to longer sequences, we conduct finetuning experiments with 2 times the max sequence length seen in pretraining (indicated by an ``ext.'' suffix after the model name). For example, ``Shatter\textsubscript{GPU} 512-$l$ ext.'' in Table~\ref{tab:glue} indicates that Shatter\textsubscript{GPU} is extended to length 512 in finetuning, while it is pretrained on sequence length 256. We compare it with (i) BERT\textsubscript{GPU} 256-$l$, which is pretrained and finetuned on sequence length 256, and (ii) BERT\textsubscript{GPU} 512-$l$ ext., which is pretrained on 256 but finetuned on 512, with the extra position embeddings randomly initialized. We can see Shatter\textsubscript{GPU} outperforming BERT\textsubscript{GPU} on most tasks, and the performance of BERT\textsubscript{GPU} decreasing when we extend sequence length, possibly due to un-pretrained position embeddings.

More systematically, in Table~\ref{tab:squad} we compare finetuned results on SQuAD v1.1 dev set, with or without extending sequence length. This question-answering task requires sequence length to be sufficiently long to cover the correct answer; hence the models with short length, BERT\textsubscript{GPU} 256-$l$ and Shatter\textsubscript{GPU} 256-$l$, suffer from low performance. When max length is extended, the performance of BERT usually decreases while {\Shatter} increases, showing the better generalizability of {\Shatter}.

Further on the MultiRC task (Table~\ref{tab:multirc}), we verify again that Shatter\textsubscript{Large} can generalize to a longer sequence length and outperform BERT\textsubscript{Large}.

\begin{table}[t!]
\footnotesize
\centering
\setlength{\tabcolsep}{12pt}
\begin{tabular}{lrr}
\toprule
BoolQ             & \multicolumn{1}{c}{Dev.} & \multicolumn{1}{c}{Test} \\ \midrule
BERT\textsubscript{Large}        & \textbf{84.5} & 82.6                       \\
Shatter\textsubscript{Large}      & 84.3 &   \textbf{82.7}             \\
BERT\textsubscript{Large} 400k   &  81.8        & 80.2                   \\
Shatter\textsubscript{Large} 400k &   82.8      &  80.1                 \\ \bottomrule
\end{tabular}
\caption{Results on the BoolQ task.}
\label{tab:boolq}
\end{table}

\subsubsection{Convergence}
\label{sec:convergence}

Next, we investigate \emph{convergence}, i.e.~how many pretrained steps are required for a model to achieve certain performance. In Figure~\ref{fig:converge}, we plot the finetuned performance of base-size models on MNLI dev sets, and large-size models on BoolQ dev, as a function of pretrained steps. {\Shatter} achieves better performance with fewer pretrained steps than BERT, and Shatter\textsubscript{Base} scores the second among all base-size models, next to BERT-RPE. The good convergence is a critical factor for Shatter\textsubscript{GPU}, as Shatter\textsubscript{GPU} only sees the same amount of tokens as Shatter\textsubscript{Base} at pretrained step 400k.

For further comparison, we show performance of large-size models at pretrained step 400k in Table~\ref{tab:glue}, and BoolQ in Table~\ref{tab:boolq}, where Shatter\textsubscript{Large} outperforms BERT\textsubscript{Large} as well. It is but noteworthy that dev and test performance do not always align; so pretraining longer seems a safe choice.

%As shown on MNLI, models that model relative position (Shatter, BERT-RPE, BERT-RAB) consistently demonstrate higher performance. Out of these, \Shatter \, is much faster per step as shown in Table~\ref{tab:pretrain_cost}. This trend holds for large models as well as shown on BoolQ.

%~\citep{clark2019boolq}. \Shatter \, performs comparably to BERT while achieving higher performance for a fewer number of pretraining steps (Figure~\ref{fig:converge})

% fast convergence, i.e., achieving better performance with less pretrain steps. 

% The second angle to computational efficiency is convergence i.e. in how many steps the model is able to achieve reasonable performance. Note that this complementary to speed described above. A model can achieve 

% \subsection{MultiRC, SQuAD, ??}

% \subsection{Extending the Maximum Length in finetuning}

% Both BERT and BERT\textsubscript{Large} were increased 

\section{Conclusion}

We have presented {\Shatter}, an efficient Transformer encoder with a novel self-attention architecture. We expect broader applications in the future.

\bibliography{anthology,custom}
\bibliographystyle{acl_natbib}

\newpage
\appendix

\begin{table*}[t]
\footnotesize
\centering
\setlength{\tabcolsep}{7pt}
\begin{tabular}{lrrrrrrrrr}
\toprule
 GLUE (Dev.)   & \multicolumn{1}{c}{CoLA} & \multicolumn{1}{c}{SST-2} & \multicolumn{1}{c}{MRPC} & \multicolumn{1}{c}{QQP} & \multicolumn{1}{c}{STS-B} & \multicolumn{1}{c}{MNLI-m} & \multicolumn{1}{c}{MNLI-mm} & \multicolumn{1}{c}{QNLI} & \multicolumn{1}{c}{RTE} \\ \midrule
Part\_Mask (\S\ref{sec:relpartition})  & 69.4                     & 81.4                      & 81.9                     & 87.2                    & 43.5                      & 80.4                       & 81.0                        & 84.7                     & 55.2                    \\
1H\_Sigmoid (\S\ref{sec:sigmoid-att}) & 79.8                     & 92.5                      & 91.3                     & 87.1                    & 83.3                      & 84.5                       & 84.7                        & \textbf{91.4}            & 61.7                    \\
Part\_Bias (\S\ref{sec:part-embeddings}) & \textbf{84.2}            & \textbf{93.8}             & \textbf{92.3}            & 85.7                    & 84.2                      & 84.5                       & 84.7                        & 91.0                     & 62.8                    \\
Shatter\textsubscript{Base}  & 84.0                     & 93.6                      & 92.1                     & \textbf{87.6}           & \textbf{85.8}             & \textbf{84.8}              & \textbf{84.9}               & 90.7                     & \textbf{67.9}           \\ \bottomrule
\end{tabular}
\caption{Finetuning results of several model variants on GLUE development set.}
\label{tab:ablation_glue}
\end{table*}

\begin{table*}[t]
\footnotesize
\centering
\setlength{\tabcolsep}{7pt}
\begin{tabular}{lrrrrrrrrr}
\toprule
  GLUE (Dev.)    & \multicolumn{1}{c}{CoLA} & \multicolumn{1}{c}{SST-2} & \multicolumn{1}{c}{MRPC} & \multicolumn{1}{c}{QQP} & \multicolumn{1}{c}{STS-B} & \multicolumn{1}{c}{MNLI-m} & \multicolumn{1}{c}{MNLI-mm} & \multicolumn{1}{c}{QNLI} & \multicolumn{1}{c}{RTE} \\ \midrule
BERT\textsubscript{Base}             & \textbf{83.6}            & 92.2                      & \textbf{89.7}            & 87.3                    & \textbf{85.6}             & \textbf{84.0}              & 83.8                        & \textbf{90.7}            & \textbf{65.3}           \\
BERT\textsubscript{Base} \texttt{[CLS]}   & 83.0                     & \textbf{93.1}             & 88.6                     & \textbf{87.9}           & 85.5                      & \textbf{84.0}              & \textbf{83.9}               & 90.5                     & 64.3                    \\ \midrule
Shatter\textsubscript{Base}           & \textbf{84.0}            & \textbf{93.6}             & \textbf{92.1}            & 87.6                    & \textbf{85.8}             & \textbf{84.8}              & \textbf{84.9}               & \textbf{90.7}            & \textbf{67.9}           \\
Shatter\textsubscript{Base} \texttt{[CLS]} & 83.5                     & 93.2                      & 90.6                     & \textbf{87.7}           & 83.9                      & 84.3                       & 84.5                        & 90.6                     & 65.3                    \\ \midrule
BERT-RPE              & \textbf{84.4}            & 93.5                      & 92.1                     & \textbf{88.5}           & \textbf{86.4}             & \textbf{86.2}              & \textbf{86.1}               & \textbf{92.3}            & \textbf{74.0}           \\
BERT-RPE \texttt{[CLS]}    & 83.7                     & \textbf{94.3}             & \textbf{92.9}            & 88.1                    & 85.6                      & 85.9                       & 85.7                        & 91.8                     & 66.4                    \\ \bottomrule
\end{tabular}
\caption{Comparing sequence classification strategy on GLUE development set.}
\label{tab:cls_strategy}
\end{table*}

\section{More Details for {\Shatter}}
\label{sec:appendix}

In this appendix, we show more detailed settings of {\Shatter}, as well as additional ablation experiments.

\subsection{Partition of Unity}
\label{sec:part_unity}

Given the number of parts $n$, the partition of unity $\{f_h\}$ $(0\leq h<n)$ consists of $n$ functions. We use half of the functions to cover relative positions to the left (i.e.~$x<0$), and the other half to the right ($x>0$).

Let $D=n/2-1$. The \textbf{Bernstein polynomials of degree $D$} consists of $D+1=n/2$ polynomials, $B_{\nu}(u)$ $(0\leq\nu\leq D)$, of the varialbe $u$. They are defined by:
\begin{align*}
B_{\nu}(u)=\binom{D}{\nu}u^{\nu}(1-u)^{D-\nu}
\end{align*}
where $\binom{D}{\nu}$ is the binomial coefficient. $\{B_{\nu}(u)\}$ is a partition of unity on the interval $0\leq u\leq 1$, and we transform the interval $u\in[0,1]$ to $x\in[0,\infty]$, in order to obtain $n/2$ functions that cover the relative positions to the right. Then, we reflex the $x$-axis to obtain another $n/2$ functions to cover the relative positions to the left.

Formally, we transform $u\in[0,1]$ to $x\in[0,\infty]$ by setting
\begin{align*}
u=\frac{\ln{\left(e^{\beta x}(1-e^{\alpha})+e^{\alpha}\right)}}{\alpha}
\end{align*}
where $\alpha<0$ and $\beta<0$ are hyper-parameters. The above is an affine transformation of the softplus; it maps $x=0$ to $u=0$, and $x=+\infty$ to $u=1$. As $\alpha\to-\infty$, the limit of $u$ converges to the hinge function $\min(\frac{\beta}{\alpha}x,1)$.

In this work, we set
\begin{align*}
\alpha&=-\frac{k+1}{L}D \\
\beta&=-\frac{1}{D}\left(\frac{D}{12}\right)^{\frac{k+1}{L}}
\end{align*}
where $0\leq k<L$ is the layer index and $L$ is the number of layers. Thus, the first layer (i.e.~$k=0$) has more parts that concentrate to relative positions close to $0$, while the last layer (i.e.~$k=L-1$) has all parts more evenly distributed with interval $\approx\!12$.

In our preliminary experiments, we have tried other settings (e.g.~all layers using the same partition), and the pretraining loss and finetuned performance are sensitive to the choice of the partition of unity. The current setting is largely based on our intuition, but might not yet be optimal.

\subsection{Pretraining Settings}
\label{sec:pretrain_setting}

\begin{figure}[t!]
\centering
\includegraphics[scale=0.48,viewport=0 0 6.3in 2.8in,clip]{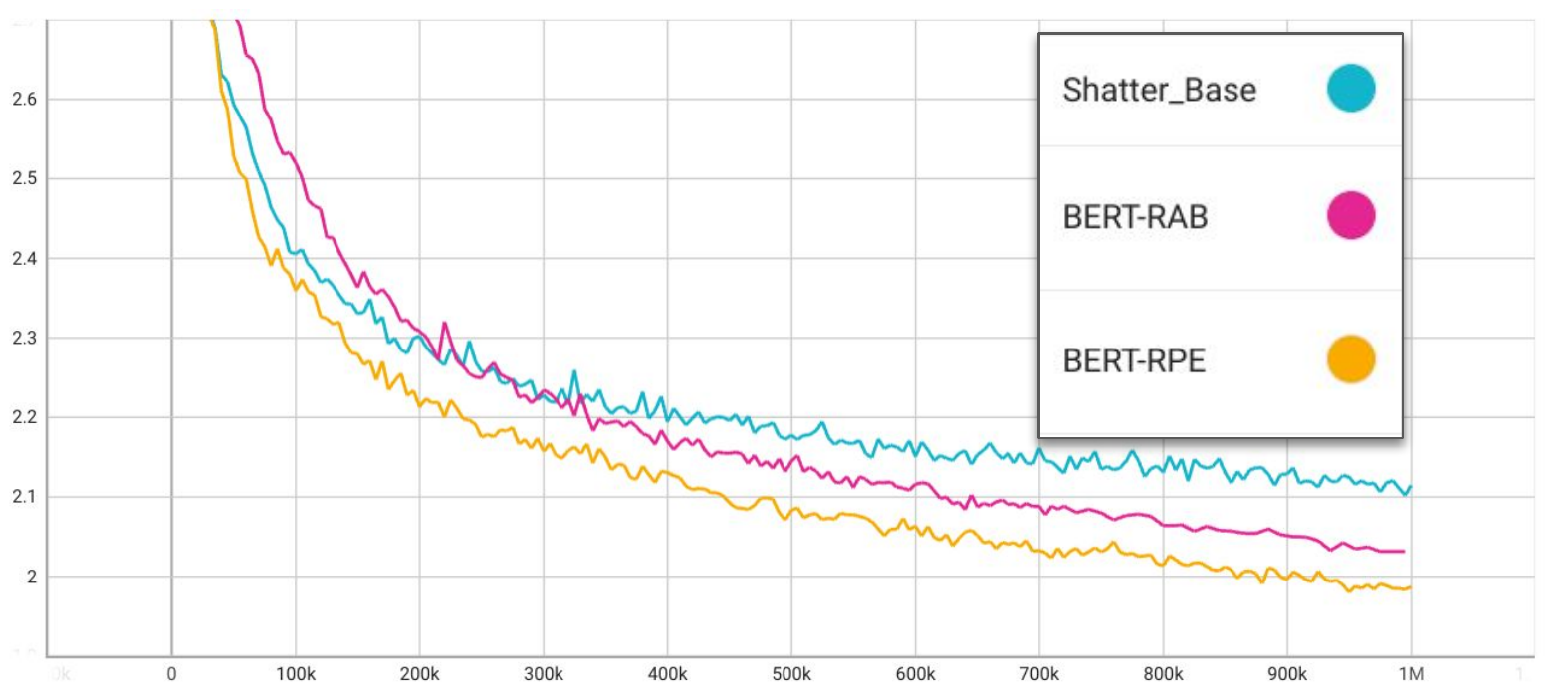}
\caption{MLM valid loss for base-size models.}
\label{fig:valid_r}
% \vspace{-0.3cm}
\end{figure}

\begin{figure}[t!]
\centering
\includegraphics[scale=0.48,viewport=0 0 6.3in 2.8in,clip]{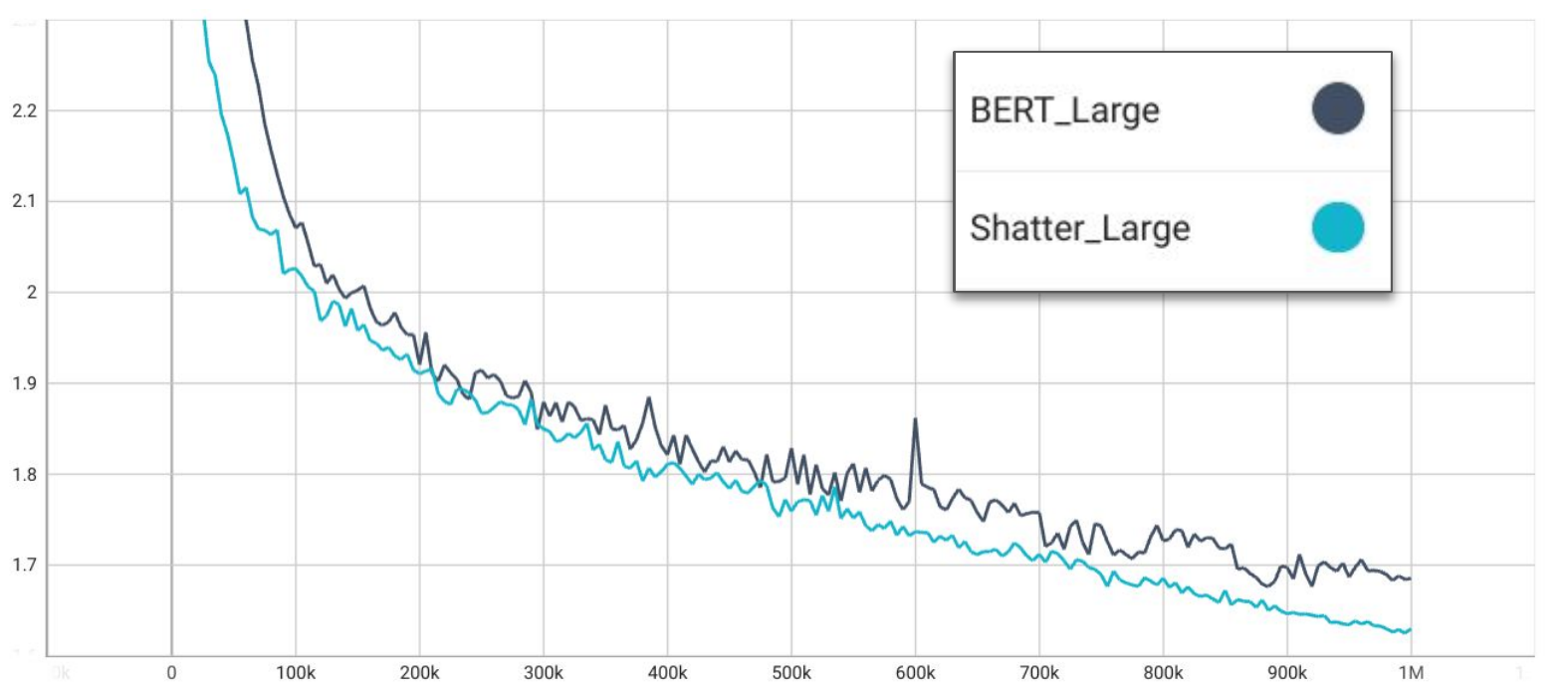}
\caption{MLM valid loss for large-size models.}
\label{fig:valid_large}
% \vspace{-0.3cm}
\end{figure}

Instead of the word-piece tokenizer~\citep{wu2016google} used in BERT, we use sentence-piece~\citep{kudo2018sentencepiece} for tokenization, on account of its easy usage and fast speed. The pretraining corpus is lower-cased and the vocabulary size set to 32k. This same tokenizer is used in all models in our experiments, for fair comparison.

To save pretraining time, we tokenize the corpus in advance and cache the results to files. The tokenization can be done with 20 parallel 4-core CPU machines in 2 hours. Following RoBERTa \citep{liu2019roberta}, we use Masked Language Modeling (MLM) as the pretraining objective, without Next Sentence Prediction.

For optimization we use Adam with $0.01$ weight decay \citep{bentivogli2009fifth}. The learning rate is set to 1e-4, with 10k steps warmup then linear decay to 0.

For validation we use the Penn Tree Bank corpus \citep{marcus-etal-1994-penn}, and the valid loss is plotted in Figure~\ref{fig:ablation} (for our model variants and BERT\textsubscript{Base}), Figure~\ref{fig:valid_r} (for base-size models) and Figure~\ref{fig:valid_large} (for large-size models). In Figure~\ref{fig:valid_r}, BERT-RPE consistently outperforms other models, while Shatter\textsubscript{Base} shows faster convergence than BERT-RAB, which roughly aligns with the convergence of finetuned performance (Figure~\ref{fig:converge}). In Figure~\ref{fig:valid_large}, Shatter\textsubscript{Large} outperforms BERT\textsubscript{Large}. The valid loss for XLNet is not shown in the figures, since XLNet is pretrained with PLM instead of MLM.

\subsection{Ablation on GLUE}
\label{sec:ablation_glue}

In Table~\ref{tab:ablation_glue}, we compare the finetuned performance of our model variants developed in \S\ref{sec:approach}. Mostly, the performance of Shatter\textsubscript{Base} $\geq$ Part\_Bias $>$ 1H\_Sigmoid $>$ Part\_Mask, which aligns with the comparison of pretraining loss in Figure~\ref{fig:ablation}. The performance of Shatter\textsubscript{Base} and Part\_Bias are close, but Shatter\textsubscript{Base} slightly outperforms Part\_Bias on MNLI and QQP.

\subsection{Sequence Classification Strategy}
\label{sec:cls_strategy}

We further compare different sequence classification strategies in Table~\ref{tab:cls_strategy}. The sequence classification method originally used in BERT \citep{devlin-etal-2019-bert}, which takes the hidden state at the \texttt{[CLS]} token, is indicated by a ``\texttt{[CLS]}'' suffix in the table. It is compared with our modified sequence classification method described in \S\ref{sec:cls_relative}.

For BERT\textsubscript{Base}, the difference in sequence classification strategy does not affect finetuned performance; but for Shatter\textsubscript{Base} and BERT-RPE, our modified method is generally better, which suggests that our method can better cope with the relative position modeling of the \texttt{[CLS]} token.

\end{document}